%% file: main.tex
\documentclass[10pt,twocolumn,letterpaper]{article}

\newif\ifarxiv
\newif\ifcvpr
\newif\ifcvprfinal

\cvprfalse \cvprfinalfalse \arxivtrue %% Arxiv
\ifcvprfinal
\usepackage{cvpr} % To produce the CAMERA-READY version
\fi
\ifcvpr
\usepackage[review]{cvpr} % To produce the REVIEW version
\fi
\ifarxiv
\usepackage[pagenumbers]{cvpr} % To produce the Arxiv version
\fi

\makeatletter
\DeclareRobustCommand\onedot{\futurelet\@let@token\@onedot}
\def\@onedot{\ifx\@let@token.\else.\null\fi\xspace}

\makeatother

\makeatletter
\renewcommand\paragraph{\@startsection{paragraph}{4}{\z@}%
  {4pt}% space before
  {-12pt}% space after
  {\normalfont\normalsize\bfseries}}
\makeatother

\expandafter\def\expandafter\normalsize\expandafter{%
    \normalsize%
    \setlength\abovedisplayskip{0pt plus 2pt}% CVPR default: 10pt plus 2pt minus 5pt
    \setlength\belowdisplayskip{0pt plus 2pt}% CVPR default: 10pt plus 2pt minus 5pt
    \setlength\abovedisplayshortskip{0pt plus 2pt}% CVPR default: 0pt plus 3pt
    \setlength\belowdisplayshortskip{0pt plus 2pt}% CVPR default: 6pt plus 2pt minus 5pt
}

\input{preamble}

\definecolor{cvprblue}{rgb}{0.21,0.49,0.74}
\usepackage[pagebackref,breaklinks,colorlinks,citecolor=cvprblue]{hyperref}
\usepackage{comment}
\definecolor{darkorange}{rgb}{1.0, 0.54, 0}

\newcommand{\coolname}{\texttt{Any\-4D}\xspace}

\newcommand{\webpagetext}{any-4d.github.io}
\newcommand{\webpage}{https://any-4d.github.io/}

\def\paperID{17783} % *** Enter the Paper ID here
\def\confName{CVPR}
\def\confYear{2026}

\definecolor{any4dpurple}{RGB}{125, 50, 200} % Cool purple for website
\definecolor{any4dblue}{RGB}{80, 140, 220}   % Cool blue for authors

\ifcvprfinal
\title{\coolname: Unified Feed-Forward Metric 4D Reconstruction \\[1.5pt]
\Large{\href{\webpage}{\color{any4dpurple}{\webpagetext}}}
\vspace{-1.2em}
}
\fi
\ifcvpr
\title{Any4D: Unified Feed-Forward Metric 4D Reconstruction}
\fi
\ifarxiv
\title{Any4D: Unified Feed-Forward Metric 4D Reconstruction \\[1.6pt]
\Large{\href{\webpage}{\color{any4dpurple}{\webpagetext}}}
\vspace{-1em}
}
\fi

\newcommand{\supptitle}[1]{
Any4D: Unified Feed-Forward Metric 4D Reconstruction
\vspace{-0.4em}
}

\definecolor{citegray}{gray}{0.32}
\newcommand{\authorhref}[3][any4dblue]{\href{#2}{\color{#1}{#3}}}
\def\authorspace{\hspace{.1in}}

\author{%
\authorhref{https://jaykarhade.github.io/}{\textbf{Jay Karhade}}
\authorspace
\authorhref{https://nik-v9.github.io/}{\textbf{Nikhil Keetha}}
\authorspace
\authorhref{https://infinity1096.github.io/}{\textbf{Yuchen Zhang}}
\authorspace
\authorhref{https://www.linkedin.com/in/tanisha-gupta-2a1934221/}{\textbf{Tanisha Gupta}}
\\[0.25em]
\authorhref{https://akashsharma02.github.io/}{\textbf{Akash Sharma}}
\authorspace
\authorhref{https://theairlab.org/team/sebastian/}{\textbf{Sebastian Scherer}}
\authorspace
\authorhref{https://www.cs.cmu.edu/~deva/}{\textbf{Deva Ramanan}}
\\[0.65em]
\href{https://www.ri.cmu.edu/}{\textcolor{citegray}{\textbf{Carnegie Mellon University}}}%
\vspace{-2em}
}

\begin{document}

% \maketitle

\twocolumn[{%
\renewcommand\twocolumn[1][]{#1}%
\maketitle
\begin{center}
    \centering
    \captionsetup{type=figure}
    \includegraphics[width=\textwidth]{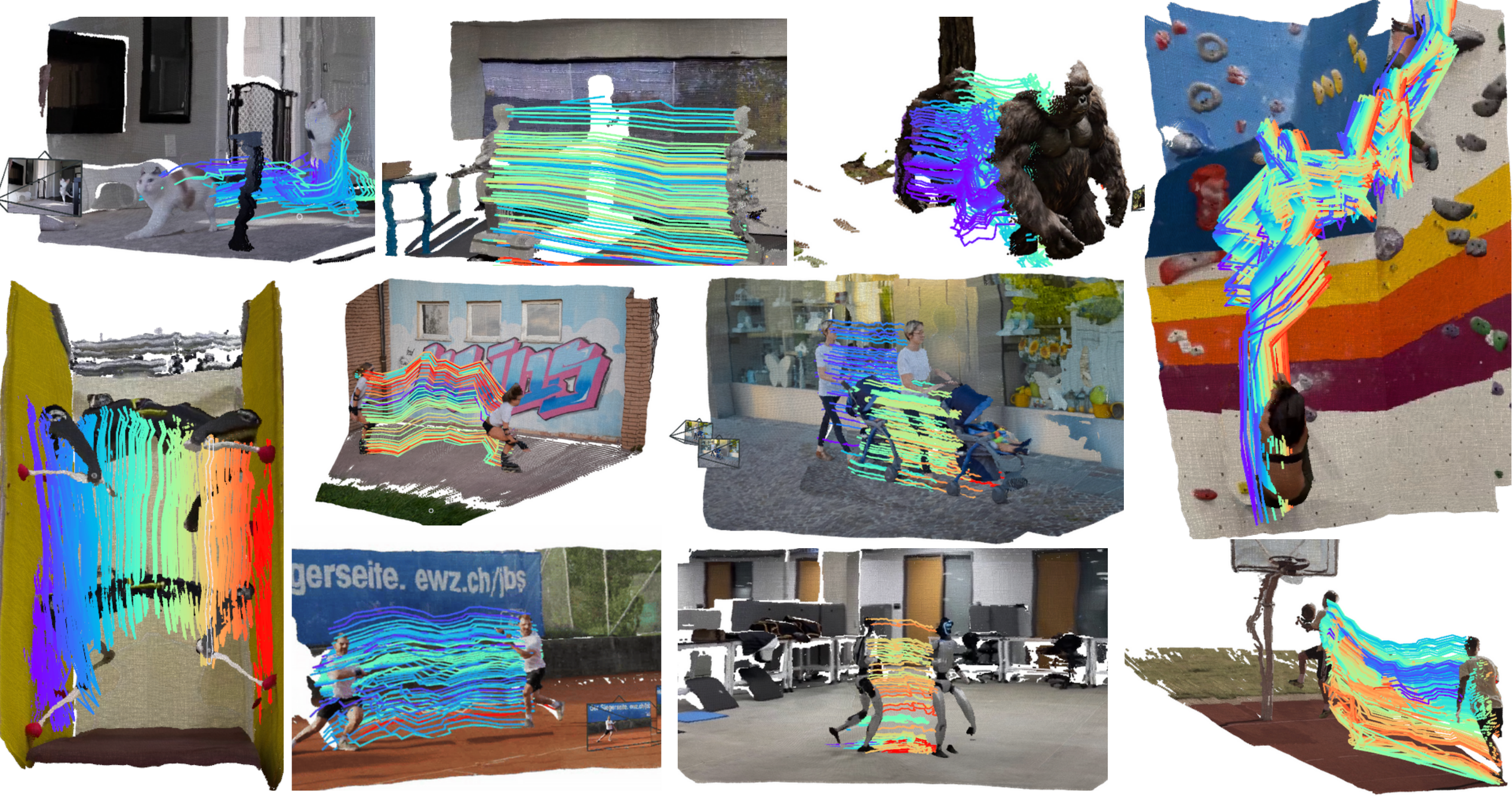}
    \captionof{figure}{\textbf{\coolname is a flexible feed-forward model capable of producing dense metric 4D reconstructions using N frames as input.} \coolname is up to $15\times$ faster and $3\times$ better than prior state-of-the-art, where performance can be further boosted by using diverse sensors as input. Note that \coolname produces dense 3D tracking vectors, but here we visualize the sparse 3D motion tracks for simplicity.}
    \label{fig:teaser}
\end{center}%
}]

\input{text/0_abstract}

\input{text/1_introduction}
\input{text/2_related_work}
\input{text/3_method}

\input{text/4_benchmarking}
\input{text/6_conclusion}

\ifarxiv
\input{supplementary}
\fi

\newpage
{
    \small
    \bibliographystyle{ieeenat_fullname}
    \bibliography{main}
}

\ifcvpr
\input{supplementary}
\fi
\ifcvprfinal
\input{supplementary}
\fi

\end{document}

%% file: preamble.tex
\usepackage[table, xcdraw, dvipsnames]{xcolor}
\usepackage{multirow}
\usepackage{graphicx}
\usepackage{float}
\usepackage{svg}
\usepackage{arydshln} %
\usepackage{titletoc} %
\usepackage{cuted}
\usepackage{amsmath}
\usepackage{amssymb}
\usepackage{bbm}

\newtoggle{comments}
\toggletrue{comments}
\iftoggle{comments}{%
    \newcommand{\akash}[1]{{\leavevmode\color{magenta}[Akash: #1]}}
    \newcommand{\deva}[1]{{\leavevmode\color{red}[Deva: #1]}}

}{%
  \newcommand{\akash}[1]{}
  \newcommand{\deva}[1]{}
  
}

\newcommand{\Lc}{\mathcal{L}}

\newcommand{\Rb}{\mathbb{R}}

\newcommand{\Gv}{\mathbf{G}}

\newcommand{\Iv}{\mathbf{I}}

\newcommand{\Ov}{\mathbf{O}}

\newcommand{\norm}[1]{\left\lVert#1\right\rVert}

\usepackage{pifont}

\definecolor{first}{rgb}{0.7, 1.0, 0.7} %
\definecolor{second}{rgb}{1, 1, 0.7} %
\definecolor{third}{rgb}{1, 0.85, 0.7} %

\newcommand{\absrel}{\textrm{rel}}
\newcommand{\threshI}{\tau}

\newcommand{\delinliers}{{\delta}_{1.25}}

\definecolor{darkgreen}{RGB}{0, 100, 0}
\newcommand{\greencheck}{{\color{darkgreen}\checkmark}}
\newcommand{\redx}{{\color{red}\ding{55}}}

%% file: text/0_abstract.tex
\begin{abstract}
\vspace{-1em}

We present \coolname, a scalable multi-view transformer for metric-scale, dense feed-forward 4D reconstruction. \coolname directly generates per-pixel motion and geometry predictions for $N$ frames, in contrast to prior work that typically focuses on either 2-view dense scene flow or sparse 3D point tracking. Moreover, unlike other recent methods for 4D reconstruction from monocular RGB videos, \coolname can process additional modalities and sensors such as RGB-D frames, IMU-based egomotion, and Radar Doppler measurements, when available. One of the key innovations that allows for such a flexible framework  is a modular representation of a 4D scene; specifically, per-view 4D predictions are encoded using a variety of egocentric factors (depthmaps and camera intrinsics) represented in local camera coordinates, and allocentric factors (camera extrinsics and scene flow) represented in global world coordinates. We achieve superior performance across diverse setups - both in terms of accuracy (\emph{$2-3\times$} lower error) and compute efficiency (\emph{$15\times$} faster) - opening avenues for multiple downstream applications.
\end{abstract}
\vspace{-1em}

%% file: text/1_introduction.tex
\section{Introduction}

Reconstructing the 4D ($3\text{D} + t$) world from sensor observations is a long-standing goal of computer vision. Such a technology can unlock a wide range of downstream tasks. In generative AI, 4D reconstruction can improve dynamic video synthesis \cite{van2024generative, lin2025movies, wu2024cat4d, chen2025reconstruct}, video understanding \cite{zhou2025llava, huang2025understanding}, and the creation of interactive dynamic assets such as VR avatars. In robotics, 4D scene reconstruction can significantly improve predictive control (MPC) for an agent navigating and manipulating in a physical world \cite{liu2025geometry, niu2025pre}. 

Although there has been significant recent progress on 4D reconstruction~\cite{zhang2024monst3r, som2024, ren2024l4gm, lei2025mosca, feng2025st4rtrack, jin2024stereo4d, lin2025movies}, dynamic reconstruction of in the wild videos remains challenging for many reasons. First, 4D reconstruction is severely under-constrained, requiring simplifying assumptions such as rigid motion, smoothness priors, or a mostly-static world assumption. Second, there is a lack of large-scale 4D datasets. Unlike million-scale video~\cite{chen2024panda} and 3D datasets \cite{avetisyan2024scenescript, antequera2020mapillary}, \textit{reliable high-quality} 4D reconstruction datasets are still limited to a few thousand scenes, primarily obtained via simulation \cite{zheng2023point, greff2021kubric}. Third, because 4D reconstruction and tracking is such a challenging problem, progress has been largely achieved by treating dynamic attribute prediction as independent sub tasks (i.e., 3D tracking, video-consistent depth estimation, scene flow estimation, camera pose estimation in dynamic scenes). This focus on sub tasks has led to fragmented datasets and benchmarks that lack consistent 4D definitions and annotations. This is unsatisfying because all sub tasks observe the same underlying 4D world!

To create a universal system that can reliably work on in the wild videos,  we seek to address the following desiderata: a) \textbf{efficiency}: much prior work often makes use of iterative optimization-based methods as a post-processing step that maybe too slow for real-time deployment. b) \textbf{multi-modality}: Many robotic platforms use additional sensors beyond cameras, but most prior work fails to exploit such diverse configurations. c) \textbf{metric scale outputs}: while existing 4D reconstruction methods produce outputs in a normalized coordinate frame, physical agents undeniably operate in the metric-scale physical world.

Taking a step in this direction, \coolname is a unified and scalable model with the following 3 core contributions:

\begin{itemize}
    \item \textbf{Dense Metric-Scale 4D Reconstruction}: \coolname predicts the dense geometry and motion of the scene in metric coordinates, unlike existing methods that can reconstruct only up-to-scale or sparse tracks. We propose a factored 4D representation consisting of per-view \emph{allocentric factors} (for scene flow and poses) and \emph{egocentric factors} (for intrinsics and depth). This factored 4D representation allows us to train on diverse datasets with partial annotations, including metric-scale 3D reconstruction datasets without motion annotations, and non-metric datasets with motion annotations.

    \vspace{0.6em}

    \item \textbf{Flexible Multi-Modal Inputs}:  When available, \coolname can further improve its 4D reconstruction by exploiting additional input modalities like depth from RGBD sensors, camera poses from IMUs or doppler velocity from RADARs compared to image-only 4D reconstruction.

    \vspace{0.6em}

    \item \textbf{Efficient Inference}:  \coolname infers both geometry and motion from  $N$ video frames in a single feed-forward pass, bypassing existing work that only predict motion for 2 frame inputs or require computationally-expensive optimization, making \coolname up to \emph{$15 \times$} faster than the next best performing method.
\end{itemize}

%% file: text/2_related_work.tex
\begin{figure*}[ht]
    \centering
    \includegraphics[width=\linewidth]{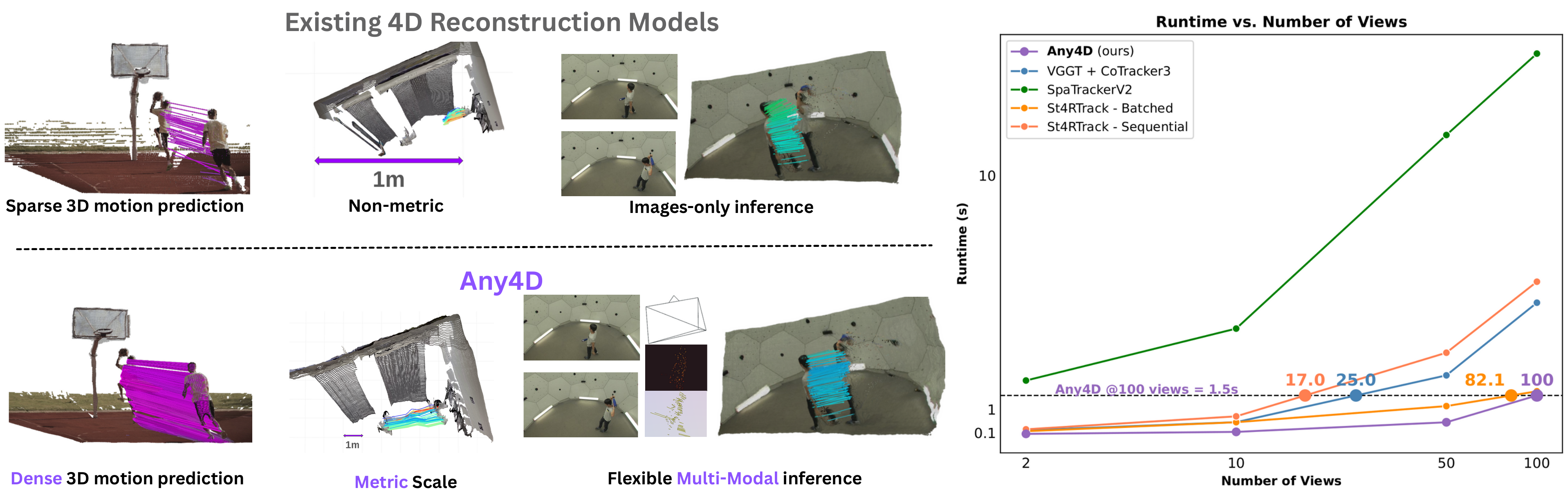}
    \caption{\label{fig:any4d_capabilities}
    \textbf{\coolname's unified capabilities overcome major limitations of existing 4D reconstruction models. }}
\end{figure*}

\vspace{-1em}

\section{Related Work}
\label{sec:related_work}

\paragraph{Reconstruction of Dynamic Scenes:} 

Reconstruction and camera pose estimation for static scenes has a rich history. It has been studied as Simultaneous Location and Mapping (SLAM)~\cite{mur2017orb, klein2007parallel, engel2014lsd, dellaert2017factor, sharma2021compositional, keetha2024splatam} when visual observations occur in a temporal sequence, and as structure-from-motion (SFM)~\cite{schonberger2016structure, agarwal2009buildingrome, seitz2006comparison, triggs1999bundle} otherwise. Since traditional optimization-based reconstruction is at odds with dynamics reconstruction, many approaches relied on ad-hoc semantic and motion masks to discard dynamic regions of a scene~\cite{qiu2022airdos, henein2020dynamic, kumar2017monocular, bescos2018dynaslam}. Subsequently, advances in data-driven monocular depth~\cite{ranftl2021vision, yang2024depth, ranftl2022midas, duisterhof2025rayst3r} and optical flow~\cite{teed2020raft, zhang2025ufm} estimation have not only enabled data-driven static reconstruction methods~\cite{teed2021droid}, but have also sparked research~\cite{li2024megasam, wu2024cat4d, kopf2021robust, lei2025mosca, seidenschwarz2025dynomo, matsuki20254dtam} in dynamic scene reconstruction. Although methods such as MegaSaM~\cite{li2024megasam} are promising, they rely on per-scene optimization, making them ill-suited for real-time use. More recently, following the success of end-to-end methods\cite{Wang_2024_CVPR, jang2025pow3r}, methods such as MonST3R~\cite{zhang2024monst3r} handle dynamic scenes by making independent per-frame predictions. However, they still require post-hoc optimization to establish explicit correspondences. To alleviate this, \cite{wang2025vggt, keetha2025mapanything, wang2025pi} also show the potential of feed-forward multi-view inference from a set of images. Following this line of work, \coolname is a feed-forward model that predicts camera poses, dense 3D motion (as scene flow) and geometry (as pointmaps), fully describing a dynamic scene captured by a set of $N$ frames in its entirety.

\paragraph{Scene Flow:} 

Scene flow was introduced in~\cite{vedula1999three} as the problem of recovering the 3D motion vector field for every point on every surface observed in a scene. Any optical flow then is the perspective projection of scene flow onto the camera plane. Subsequently, it has been studied through a wide range of approaches, ranging from variational methods~\cite{huguet2007variational, basha2013multi, pons2007multi} to learning-based supervised methods ~\cite{liu2019flownet3d, wang2019dynamic} and self-supervised methods~\cite{wu2020pointpwc, mittal2020just, puy20flot}. Despite these advances, solutions to scene flow estimation have largely been tailored to specific downstream use cases, exploiting access to privileged information. In autonomous vehicles (AVs), scene flow approaches \cite{vedder2024neural, Chodosh_2024_WACV} typically access sensor pose through inertial and proprioceptive sensors. Similarly, RAFT-3D~\cite{teed2021raft3d} assumes access to depth. Recently, \cite{lin2025movies} proposed to build upon \cite{wang2025vggt} for scene flow and view synthesis. However, in the spectrum of dynamism in a scene, we observe that all the above scene flow methods are limited to simplistic scenes like \cite{cabon2020virtual, mehl2023spring, Menze2015CVPR} with minimal dynamic motion. Our model is instead capable of directly predicting scene flow in the \emph{allocentric} coordinate frame .

\paragraph{3D Tracking:} 

While scene flow has been defined for short-range motion typically for a pair of image frames, point tracking~\cite{1641022} is the task of tracking a pixel trajectory over a long time horizon. Following methods such as~\cite{harley2022particle, karaev23cotracker, karaev2024cotracker3} that show the success of 2D point-tracking, TAPVID-3D~\cite{koppula2024tapvid3d} introduced a benchmark to address the problem of 3D point tracking. Subsequently, \cite{xiao2024spatialtracker, zhang2025tapip3d, ngo2024delta} proposed methods for obtaining 3D point tracks and improving this benchmark. However, \cite{xiao2024spatialtracker, ngo2024delta} focus on ego-centric 3D point-tracking, unlike \coolname which regresses \emph{allocentric} 3D point-tracks. \cite{xiao2025spatialtracker} recently proposed a method for allocentric 3D point tracking, by jointly optimizing camera motion, 2D and 3D point tracks. However, it is important to note that these methods can only track sparse points and require knowledge of poses and depth, either from ground truth or from running off the shelf models, limiting real-time deployment. In contrast, \coolname natively supports dense 3D point tracking and can take flexible inputs, allowing adoption on a range of platforms.

\paragraph{Concurrent and Recent Work:}

We acknowledge concurrent works \cite{feng2025st4rtrack, sucar2025dynamic, zhang2025pomato, han2025d, lin2025movies, jin2024stereo4d, liu2025trace} that focus on predicting geometry and motion, with \cite{jin2024stereo4d, lin2025movies} being limited to extremely small camera and scene motion. \coolname differs from all concurrent methods in 3 ways (see \cref{fig:any4d_capabilities}).  \textbf{First}, all concurrent methods require multiple feedforward passes to infer the motion, whereas \coolname adopts a scalable architecture inspired by \cite{keetha2025mapanything} and performs a single feedforward pass for all image frames at once. \textbf{Second}, these methods only accept image inputs, while \coolname which can exploit diverse multi-modal inputs. \textbf{Third}, unlike the concurrent works, we are the only method to produce metric scale 4D reconstructions. 
We believe that the open-source release of \coolname will set a strong foundation for the community.

%% file: text/3_method.tex
\begin{figure*}[ht]
    \centering
    \includegraphics[width=0.95\linewidth]{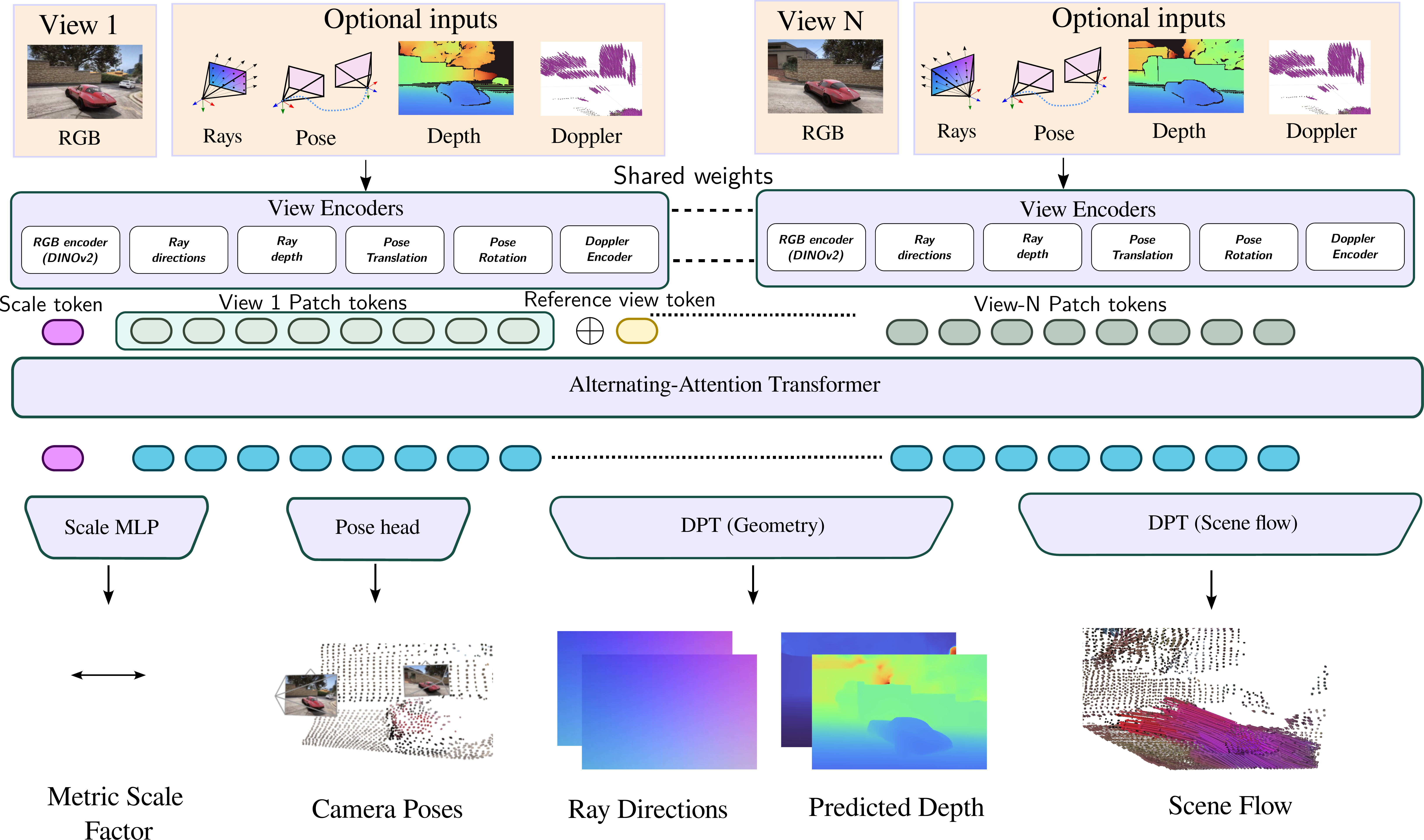}
    \caption{\textbf{\coolname predicts a factorized dense metric 4D reconstruction} represented as a global metric scale, per-view egocentric factors (depth maps and ray directions) and per-view allocentric factors (forward scene flow and camera poses) as explained in \cref{sec:method}. \coolname is a N-view transformer, consisting of modality-specific encoders, followed by an alternating-attention transformer to produce contextual patch embeddings. The output tokens from the transformer are then decoded using individual decoders specific to each factor. 
    \label{fig:architecture}}
\end{figure*}

\begin{figure*}[!ht]
    \centering
    \includegraphics[width=0.95\linewidth]{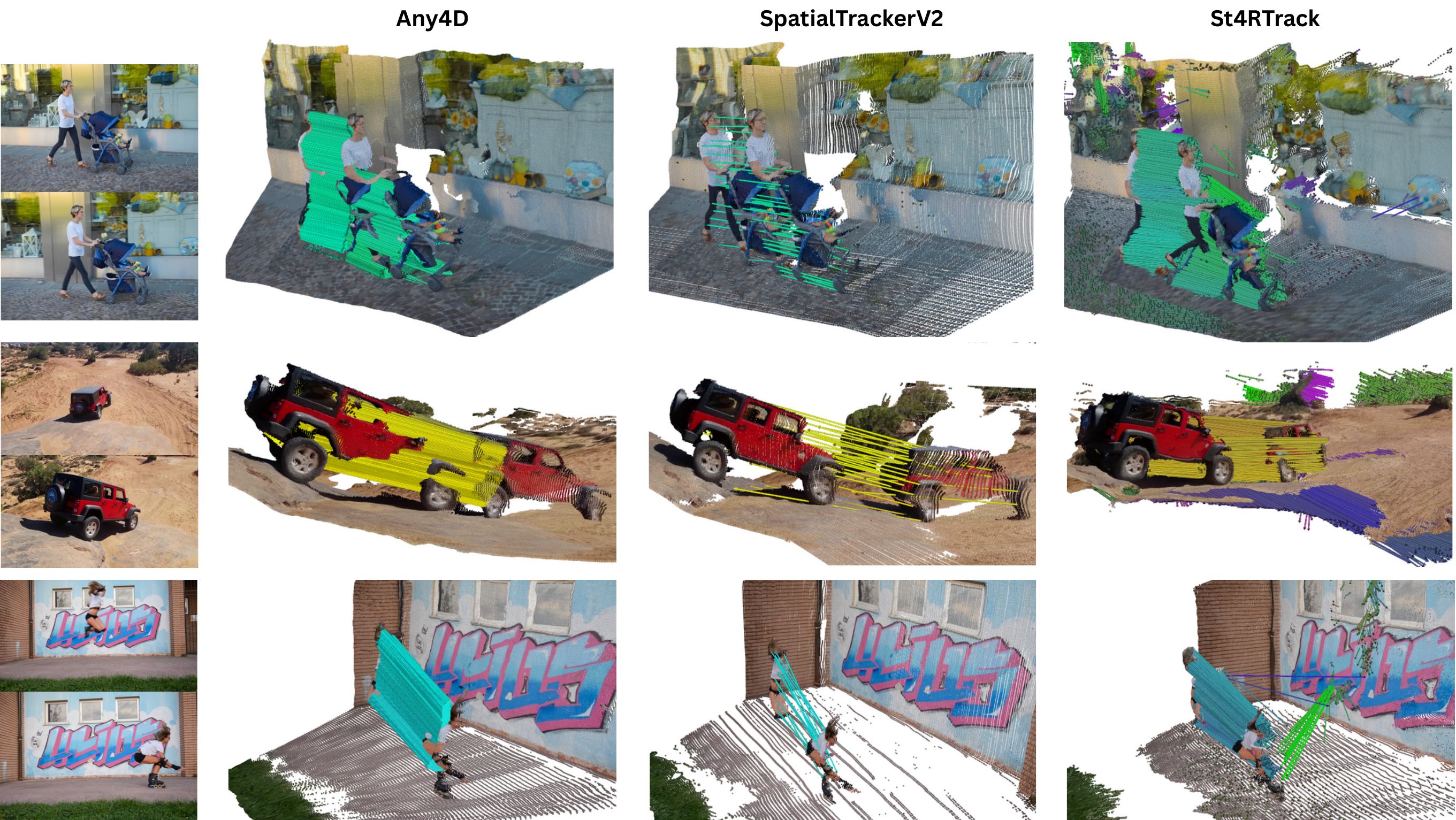}
    \caption{\label{fig:scene_flow_baseline_qual_comparison}
    \textbf{\coolname provides dense and precise motion estimation, where on the other hand, state-of-the-art baselines either produce reliable but sparse motion ({\footnotesize{SpatialTrackerV2~\cite{xiao2025spatialtracker}}}) or dense per-pixel motion that is not accurate ({\footnotesize{St4RTrack~\cite{feng2025st4rtrack}}})}.
    For SpatialTrackerV2, we are only able to uniformly query a maximum of 2500 points with a H100 GPU using 80 gigabytes of GPU memory.
    Note that we don't use any pre-computed segmentation mask but purely threshold our scene flow output to get a binary motion mask. St4RTrack cannot produce good binary motion masks due to incorrect scene flow predictions on object boundaries and the background.
    }
\end{figure*}

\section{Any4D} 
\label{sec:method}

\coolname is a transformer that takes flexible multi-modal inputs and outputs a dense metric-scale 4D reconstruction in a single feed-forward pass.
In addition to a set RGB images $\Iv \triangleq \{I_i\}_{i=1}^N$, \coolname can use auxiliary multi-modal sensor inputs which we denote as $\Ov \triangleq (O_i)_{i=1}^N$. Then, our model can be represented as a function that maps these inputs to a factored output representation as follows:
\begin{equation}
    \label{eq:any4d}
    (\tilde{s}, \{\tilde{R}_i, \tilde{D}_i, \tilde{T}_i, \tilde{F}_{i}\}_{i=1}^N) = 
    \text{Any4D}\bigl(\Iv, \Ov \bigr)
     \text{,}
\end{equation}
\noindent where the optional inputs $\Ov$ can contain information such as depth maps, camera intrinsics, camera poses from external systems or IMU and Doppler velocity from RADAR.  

Model predictions are denoted with $\sim$ in order to differentiate them from ground-truth targets or auxiliary inputs. Predictions include a metric scaling factor $\tilde{s} \in \Rb$ for the entire scene, egocentric quantities predicted in the local camera coordinate frame, namely
\begin{itemize}
    \item ray directions for each view, i.e., $\tilde{R}_i \in \Rb^{3 \times H \times W}$ 
    \item scale-normalized depth along the rays for each view, i.e., $\tilde{D}_i \in \Rb^{1 \times H \times W}$.
\end{itemize}
and \emph{allocentric} quantities predicted in a consistent world coordinate frame, namely
\begin{itemize}
   \item scale-normalized forward scene flow from the first view to all other views, i.e., $\tilde{F}_i \in \Rb^{3 \times H \times W}$.
   \item camera pose of each view in the coordinate system of the first view, i.e., $\tilde{T}_i \triangleq [p_i, q_i] \in \mathbb{R}^{7}$ represented using a scale-normalized translation vector and quaternion.
\end{itemize}

\noindent Now, given these output factors from \coolname, one can recover the predicted metric-scale geometry $\tilde{\Gv}_i$ in the form of pointmaps~\cite{Wang_2024_CVPR} by composing the individual quantities as
\begin{equation}
    \tilde{\Gv}_i = \tilde{s} \cdot \tilde{T}_i \cdot \tilde{R}_i \cdot \tilde{D}_i~~~~\in \mathbb{R}^{3 \times H \times W}.
\end{equation}
Similarly, allocentric scene flow $\tilde{M}_i$ and pointmaps after motion $\tilde{\Gv'}_i$ can be recovered as
\begin{equation}
    \tilde{M}_i = \tilde{s} \cdot \tilde{F}_i~~~~\in \mathbb{R}^{3 \times H \times W}
\end{equation}
\begin{equation}
    \tilde{\Gv'}_i = \tilde{\Gv}_i + \tilde{M}_i ~~~~\in \mathbb{R}^{3 \times H \times W}
\end{equation}
We show in \cref{sec:benchmarking}, that this parameterization of motion and geometry is optimal for model performance compared to other parameterizations.

\subsection{Architecture} \label{subsec:architecture}
\coolname largely follows a multi-view transformer architecture, similar to \cite{keetha2025mapanything} (see \cref{fig:architecture}). 
Conceptually, it can be separated into three sections: a) modality specific input encoders, b) a multi-view transformer backbone that attends to the tokens from all views, and c) output representation heads which decode the tokens into the factorized output variables for each view. 

\paragraph{Multi-Modal Input Encoders:} RGB inputs $\Iv$ and auxiliary multi-modal sensor inputs $\Ov$ are mapped to view-specific patch tokens through multi-modal view encoders with shared weights for input views which map to a $\mathbb{R}^{1024 \times H/14 \times W/14}$ feature space. We follow the design choices in \cite{keetha2025mapanything} for RGB, depth, camera poses and intrinsics encoders, and additionally, add a CNN encoder to encode doppler velocity. We summarize these below:
\begin{itemize}
    \item \textbf{RGB Images}: DINOv2~\cite{oquab2023dinov2} for encoding images, to extract the layer-normalized patch-level features from the final layer of DINOv2 ViT-Large, $F_\text{I} \in \mathbb{R}^{1024 \times H/14 \times W/14}$. 
    \item \textbf{Depth Images:} A shallow CNN encoder is used to encode depth images, where we normalize the input depth before passing it to the depth encoder. The normalization factor is computed independently for each local view.
    \item \textbf{Doppler Velocity:} Doppler velocity is also encoded using a CNN-based encoder. However, here the normalization factor for encoding the doppler velocity is computed from the first-view pointmap and shared globally.
    \item \textbf{Camera Intrinsics:} Camera intrinsics are encoded as rays, and also use a CNN that maps the 3-channel ray-directions into the same 1024-dimensional latent space.
    \item \textbf{Camera Poses:} Two 4-layer MLP encoders are used for camera rotation and translation that map normalized input poses to latent vectors, $f_\text{rot} \in \mathbb{R}^{1024}$ and $f_\text{trans} \in \mathbb{R}^{1024}$. The normalization factor for pose translation is computed globally across all views, and a positional encoding is used to indicate the reference view $p_\text{ref} \in \mathbb{R}^{1024}$. 
    \item \textbf{Metric Scale Token}: For metric-scale data, the depth scale and pose scale obtained from normalizing depth and pose are first transformed to log-scale and then encoded using a 4-layer MLP, yielding two $\mathbb{R}^{1024}$ latent features.
\end{itemize}
All multimodal encodings thus obtained are aggregated via \textit{summation} into a per-view embedding $F_{view} \in \mathbb{R}^{1024 \times H/14 \times W/14}$, which are flattened into tokens, along with an added learnable token to learn the metric-scale.

\paragraph{Transformer Backbone:} We use an alternating-attention transformer \cite{wang2025vggt} across the views, consisting of 12 blocks of 12 multi-head attention and MLPs. Each transformer block processes tokens with a latent dimension of 768 and contains MLPs with a ratio of 4, similar to the ViT-Base architecture. Furthermore, consistent with \cite{keetha2025mapanything} we choose to not use 2-D rotary positional encoding (RoPE) for the inputs, and also employ Flash Attention~\cite{dao2023flashattention} for efficiency.

\paragraph{Output Representation Heads:} We decode the multi-view tokens from the transformer backbone into a factored output representation as follows:
\begin{itemize}
    \item \textbf{Geometry DPT Head}: We use a dense prediction transformer (DPT) \cite{ranftl2021vision} head to predict per-view ray directions $\tilde{R}_i$, up-to-scale ray depths $\tilde{D}_i$, and confidence masks. 
    \item \textbf{Motion DPT Head}: A second DPT head is tasked to predict per-view forward \emph{allocentric} scene flow $\tilde{F}_i$ . The scene flow represents motion of points in the reference view-0 to all other views.
    \item \textbf{Pose Decoder}: The pose decoder is an average-pooling-based CNN decoder that predicts per-view, up-to-scale translations and quaternions $\tilde{T}_i \triangleq [p_i, q_i]$.
    \item \textbf{Metric Scale Decoder}: We use a lightweight MLP decoder to predict the log scale metric scaling factor, which is subsequently exponentiated.
\end{itemize}

\subsection{Training Details} \label{subsec:training}

\paragraph{Datasets:} Despite recent efforts~\cite{jin2024stereo4d}, there is a lack of large-scale datasets that contain dynamic scene motion annotations. In fact, \textit{reliable, high-quality} scene flow annotations are sparse and only available from simulation engines \cite{karaev2023dynamicstereo, greff2021kubric}. We address this challenge in this work by a) finetuning large-scale pretrained geometry models and b) training with partial supervision. Owing to our factored representation, we are able to train on a mixture of both geometry-only and dynamic datasets, where they can be synthetic or real-world with varying sparsity of labels: BlendedMVS~\cite{yao2020blendedmvs}, MegaDepth~\cite{li2018megadepth}, ScanNet++~\cite{yeshwanth2023scannet++}, VKITTI2~\cite{cabon2020virtual}, ParallelDomain4D~\cite{vanhoorick2024gcd}, Waymo-DriveTrack~\cite{balasingam2024drivetrack}, 
SAIL-VOS3D ~\cite{hu2021sail}
PointOdyssey~\cite{zheng2023point}, Dynamic Replica~\cite{karaev2023dynamicstereo} and Kubric \cite{greff2021kubric} data generated by CoTracker3\cite{karaev2024cotracker3} and GCD\cite{vanhoorick2024gcd}. Detailed information of all datasets used for training is available in the appendix. %

\paragraph{Training with Multi-Modal Conditioning:} We preprocess the datasets and generate multi-modal inputs offline for faster training. Geometric inputs consisting of poses, depths and intrinsics are directly taken from the dataset annotations. To simulate doppler velocity, we take the radial component of egocentric scene flow between data pairs. During training, multi-modal conditioning is applied with a probability of 0.7, i.e., 70\% of training iterations include multi-modal inputs alongside images. Additionally, we ensure that individual modalities (depth, rays, poses, and doppler) are independently removed with a probability of 0.5 to promote effective learning in flexible input configurations. Finally, we initialize our network with MapAnything weights \cite{keetha2025mapanything}. For each training batch, we sample up to 4 views from the datasets and train on 1 H100 node for 100 epochs.

\paragraph{Losses:} \coolname is trained using a combination of geometric and motion losses based on the type of annotation available. Ray directions representing the camera intrinsics and quaternions are scale-agnostic, and therefore can be supervised via simple regression losses: \begin{equation}
   \Lc_{\text{rays}} \triangleq \sum_{i=1}^N \| R_i - \tilde{R}_i \|
\end{equation}
\begin{equation}
  \Lc_{\text{rotation}} \triangleq \sum_{i=1}^N \min(\|q_i - \tilde{q}_i \|, \| -q_i + \tilde{q}_i \|). 
\end{equation}

On the other hand, geometric quantities such as camera translations $t_i$, ray depths $D_i$ and scene flow $F_i$ are predicted in a scale-normalized coordinate frame. %
Following prior work~\cite{Wang_2024_CVPR, leroy2024grounding, keetha2025mapanything}, we use the ground-truth validity masks $V_i$ and pointmaps $X_i$ and compute the ground-truth scale as the average euclidean distance of valid points with respect to the world origin (given by the first view camera frame): $z = \|\{X_i[V_i]\}_i^N\| / \sum_i^N V_i$. %
To compute scale-invariant losses, we also compute a scale factor derived from our predictions $\tilde{z} = \|\{\tilde{X}_i[V_i]\}_i^N\| / \sum_i^N V_i$:
\begin{equation}
   \Lc_{\text{trans}} \triangleq \sum_i^N \norm{ \frac{t_i}{z_i} - \frac{\tilde{t}_i} {\tilde{z}_i}},
\end{equation}
\begin{equation}
   \Lc_{depth} \triangleq \sum_i^N \norm{f_\text{log}\left(\frac{D_i}{z_i}\right) - f_\text{log}\left(\frac{\tilde{D}_i}{\tilde{z}_i}\right)}
\end{equation}
where $f_\text{log}(x) \triangleq ({ x / \| x \|})\log(1 + \| x \|)$ converts quantities to log-space for numerical stability.
A pointmap loss is also applied to the composed geometric predictions as follows:
\begin{equation}
   \Lc_{pm} \triangleq \sum_i^N \norm{f_\text{log}\left(\frac{X_i}{z_i}\right) - f_\text{log}\left(\frac{\tilde{X}_i}{\tilde{z}_i}\right)}
\end{equation}

Similarly, scene flow is also supervised in a scale-invariant manner. We find that scene flow loss is dominated by static points since most of the scene is static. Therefore, we find it is crucial to calculate a static-dynamic motion mask $M$ from the ground truth scene flow, and upweight the scene flow loss in the dynamic regions by 10x more compared to static regions:
\begin{equation}
    \Lc_{sf} \triangleq \sum_i^NM\cdot\norm{ f_\text{log}\left(\frac{F_i}{z_i}\right) - f_\text{log}\left(\frac{\tilde{F}_i} {\tilde{z}_i}\right)}
\end{equation}
Finally, the predicted metric scale factor $\tilde{s}$ is also supervised in the log space as follows: $\Lc_\text{scale} \triangleq \|f_\text{log}(z) - f_\text{log}(\tilde{s} \cdot \text{sg}(\tilde{z})\|$, where $\text{sg}$ denotes the stop-gradient operation and prevents the scale supervision from affecting other predicted quantities. The final loss is expressed as: 
\begin{equation}
    \Lc = \Lc_\text{trans} + \Lc_\text{rot} + \Lc_\text{rays} + \Lc_\text{depth} + \Lc_\text{sf} + \Lc_\text{mask}
\end{equation}

%% file: text/4_benchmarking.tex
\input{tables/3d_tracking}

\input{tables/dense_sf}

\input{tables/video_depth}

\begin{figure*}[!ht]
    \centering
    \includegraphics[width=0.95\linewidth,clip=true,trim=0mm 0mm 0mm 0mm]{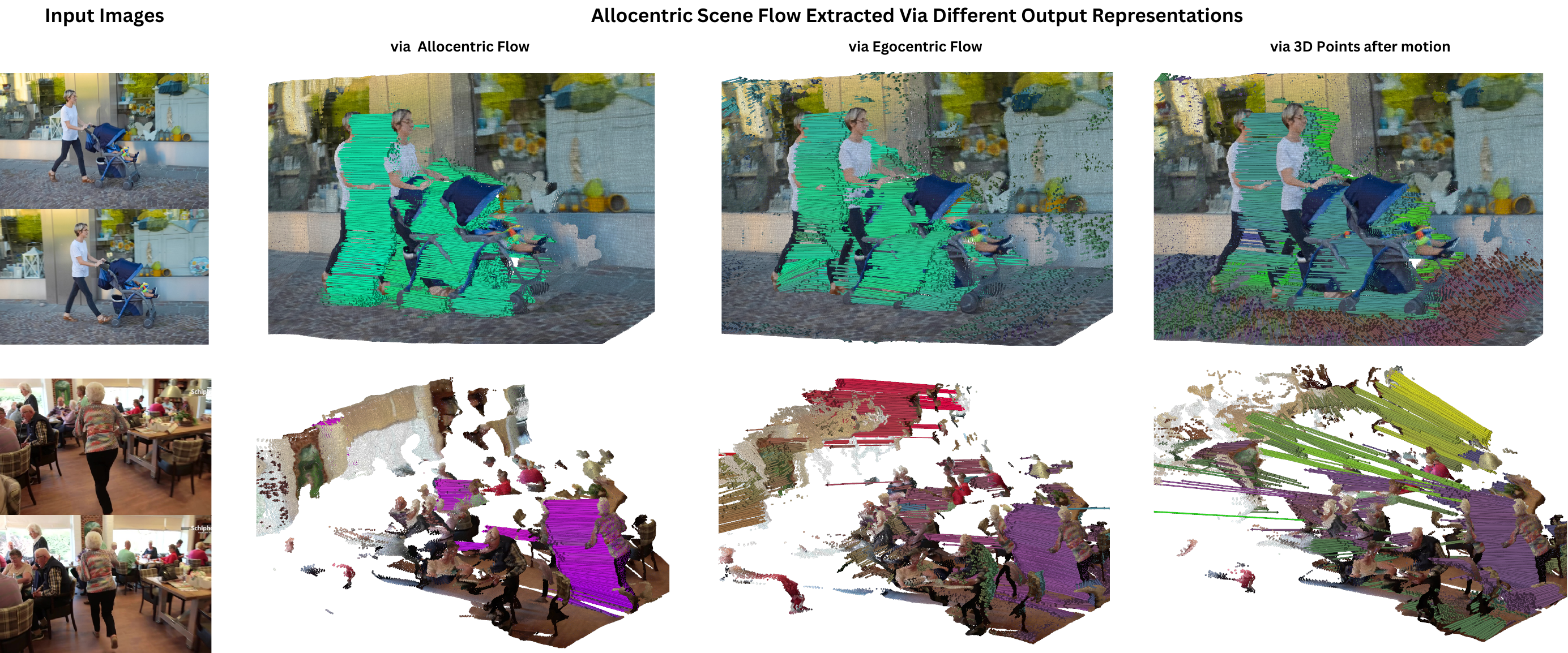}
    \caption{\label{fig:scene_flow_parameterization_ablation}%
    \textbf{Scene motion parametrized as \emph{allocentric} scene flow provides the cleanest 4D reconstructions.} We find that other parameterizations such as 3D points after motion (proposed in St4RTrack \cite{feng2025st4rtrack}) provide extreme noise on object boundaries and background.
    }
\end{figure*}

\input{tables/multi_modal}

\section{Results \& Analysis}
\label{sec:benchmarking}

We evaluate \coolname on diverse benchmarking setups specifically designed for allocentric 4D reconstruction, and compare against state-of-the-art (SOTA) methods. 

\paragraph{3D Tracking:} 

There is a lack of standard and unified benchmarks for evaluating 4D reconstruction in the existing literature. To create allocentric 3D tracking benchmarks, we follow \cite{feng2025st4rtrack} and repurpose existing 3D tracking benchmark, particularly TAPVID-3D \cite{koppula2024tapvid3d}. However, TAPVID-3D have their own limitations: the Aria Digital Twin (ADT) sequences are largely static, Parallel-Studio sequences contain fixed-camera viewpoints, while the DriveTrack sequences are extremely sparse. Hence, we choose to drop ADT, and keep Parallel-Studio and Drive-Track benchmarking sequences. We also add unseen held-out sequences from Dynamic Replica \cite{karaev2023dynamicstereo} and a zero-shot dataset LSFOdyssey \cite{wang2025scenetracker}, both of which contain camera motion along with 3D tracking labels. The final benchmark contains $\sim170$ sequences across 4 datasets of up to 64 frames in length. We evaluate \coolname against SOTA 3D trackers SpatialTrackerV2 \cite{xiao2025spatialtracker} and St4RTrack \cite{feng2025st4rtrack}. We also compose 3D reconstruction models  \cite{zhang2024monst3r, leroy2024grounding, wang2025vggt, keetha2025mapanything} with 2D tracks from CoTracker3 \cite{karaev2024cotracker3} for comparison.  

We use standard benchmarking protocols \cite{feng2025st4rtrack, xiao2025spatialtracker, teed2021raft3d, koppula2024tapvid3d} to evaluate the quality of our 4D reconstruction. Following \cite{feng2025st4rtrack}, we first perform median-scaling to align to metric space. We report average percent of points within delta for 3D points after motion ($\text{APD}$) and inlier percentage $\threshI$ for scene flow. We also report End Point Error (EPE) for 3D points after motion (dynamic points) and 3D scene flow vectors. $\text{APD}$ and $\threshI$ are defined as:
\begin{align}
    \text{APD} = \sum_{i,t} \mathbbm{1} \cdot \left(\norm{P_{i,t} - \tilde{P}_{i,t}} < \delta_{\text{3D}}\right) \\
    \threshI = \sum_{i,t} \mathbbm{1} \cdot \left(\norm{F_{i,t} - \tilde{F}_{i,t}} < 0.1\text{m}\right)
\end{align}
where $\Tilde{P_i}$ represents the predicted 3D point after motion and $\Tilde{F_i}$ is the corresponding scene flow vector at time $t$.
For APD, we use thresholds $\delta_{3D} \in \{0.1, 0.3, 0.5, 1.0\}$ m. 
As evident in \cref{tab:3d_tracking}, \coolname shows state-of-the-art performance across all datasets. Furthermore, it is $15\times$ faster than the closest performing method, SpatialTrackerV2. 
This is further reinforced qualitatively in \cref{fig:scene_flow_baseline_qual_comparison}.

\paragraph{Dense Scene Flow:}
We construct allocentric scene flow benchmarks by repurposing 2 egocentric scene flow benchmarking datasets: VKITTI-2 \cite{cabon2020virtual} and Kubric-4D \cite{van2024generative}. While scene flow in VKITTI-2 is limited to small consecutive frame motion, we can simulate scene flow across 60 frames and 16 camera viewpoints from Kubric4D (GCD). Hence we create 2 variants for Kubric4D (GCD): (a) scene flow from static camera movement and (b) scene flow from wide-baseline dynamic camera movement. Importantly, all pairs from both datasets are from held-out scenes to ensure there is no data leak from the training datasets. We evaluate \coolname against St4RTrack which can predict dense scene flow, and 3D reconstruction method outputs composed with optical flow from SEA-RAFT \cite{wang2024sea}, to calculate \textit{covisible} scene flow. We are unable to run SpatialTrackerV2 or CoTracker3 as they do not support per-pixel point queries and run out-of-memory(OOM). From \cref{tab:dense_sf}, we see that \coolname outperforms baselines by $2-3\times$ on average on $\text{APD}$, and by even more on scene-flow metrics.

\paragraph{Video Depth:}
We also evaluate \coolname on standard video depth benchmarks \cite{geiger2013vision, mayer2016large, palazzolo2019refusion} in \cref{tab:video_depth}, against specialized video depth baselines \cite{chen2025video, hu2025depthcrafter}, feed-forward + iterative optimization baselines \cite{Wang_2024_CVPR, zhang2024monst3r, li2024megasam, xiao2025spatialtracker}, and single-step feed-forward baselines \cite{wang2025continuous, wang2025vggt, keetha2025mapanything}. \coolname shows state of the art video depth estimation over other single-shot feed-forward inference baselines while being competitive with optimization based and task-specific methods.

\paragraph{Support for Multi-Modal Inputs:}
Since \coolname can utilize flexible inputs for inference to enhance performance, we study improvements to scene flow on the dense Kubric-4D static benchmark and 3D tracking on LSFOdyssey benchmark by incorporating different input modalities. From \cref{tab:multi_modal}, we observe that adding geometry significantly improves $\text{APD}$ and EPE for 3D points. Adding doppler further improves scene-flow, with the best performance achieved when all modalities are provided.

\paragraph{Choice of Motion Representation:}
While allocentric motion $F_\text{allo}$ is arguably the useful quantity for downstream applications, it is possible to represent the predicted scene flow output in 4 ways:
\begin{itemize}
    \item \textbf{Allocentric Scene Flow:} Directly predicting $\tilde{F}_\text{allo}$.
    \item \textbf{Egocentric Scene Flow:} Predicting egocentric scene flow $\tilde{F}_\text{ego}$, and using estimated geometry to recover allocentric motion as: \[
\tilde{F}_\text{allo}
= T_{t\!\to t+1}\!\bigl(P_{0}^{v} + \tilde{F}_{\text{ego}}\bigr) - p
\]
    \item \textbf{3D Points After Motion:} Predicting view-aligned pointmaps at time 0 and t - $P_{0}^{v}$ and $P_{t}^{v}$, and recovering the allocentric motion: \[
    \tilde{F}_{\text{allo}} = P_{t}^{v} - P_{0}^{v} 
    \]
    \item \textbf{Backprojected 2D Flow:} Unprojecting optical flow to obtain covisible scene flow between pointmaps.
\end{itemize}
We systematically investigate these choices in \cref{tab:representation_ablation} and \cref{fig:scene_flow_parameterization_ablation}. We find that directly predicting allocentric motion leads to optimal performance not only on \textit{scene flow}, but surprisingly, also on \textit{dynamic pointmaps after motion}, compared to directly predicting points after motion as adopted otherwise in \cite{feng2025st4rtrack}.

\input{tables/representation_ablation}

\paragraph{Limitations:}

Although \coolname takes a step forward towards achieving 4D reconstruction models, we identify important limitations.  Firstly, we always calculate scene-flow from the reference (first) view to all other frames in the sequence, necessitating that the object of interest should be present at the start of the video. One possible way to alleviate this is by training \coolname in a permutation invariant manner as in \cite{wang2025pi}. Secondly, we assume perfectly simulated multi-modal input and do not account for sensor noise - which is hardly true for real-world deployment. Finally, as with all data-driven architectures, generalization is a function of the diversity and size of the training set. We believe that \coolname's performance on highly dynamic scenes and wide baselines (or low frame-rate videos) can be improved with the availability of richer dynamic 3D datasets\cite{tesch2025bedlam2}.

%% file: tables/3d_tracking.tex
\begin{table*}[t]
\caption{\label{tab:3d_tracking}%
  \textbf{\coolname{} showcases state-of-the-art sparse 3D point tracking, while providing dense motion predictions and being an order of magnitude faster than the closest performing baseline.} We report end-point error (EPE), average points within delta (APD) and inlier ratio at 0.1m ($\threshI$) for dynamic points in the benchmark. The runtime is computed on a H100 using 50 frames as input. Best results are \textbf{bold}.
}
\centering
\scriptsize
\setlength\tabcolsep{4pt}
\resizebox{\textwidth}{!}{%
\begin{tabular}{lrrrrrrrrrrrrrrrrr}
\toprule
& & \multicolumn{4}{c}{\textbf{Drive Track}~\cite{balasingam2024drivetrack}} & \multicolumn{4}{c}{\textbf{Dynamic Replica}~\cite{karaev2023dynamicstereo}} & \multicolumn{4}{c}{\textbf{LSFOdyssey}~\cite{wang2025scenetracker}} & \multicolumn{4}{c}{\textbf{PStudio}~\cite{koppula2024tapvid3d}} \\
\cmidrule(lr){3-6} \cmidrule(lr){7-10} \cmidrule(lr){11-14} \cmidrule(lr){15-18}
& & \multicolumn{2}{c}{\textbf{Dynamic Points}}
& \multicolumn{2}{c}{\textbf{Scene Flow}}
& \multicolumn{2}{c}{\textbf{Dynamic Points}}
& \multicolumn{2}{c}{\textbf{Scene Flow}}
& \multicolumn{2}{c}{\textbf{Dynamic Points}}
& \multicolumn{2}{c}{\textbf{Scene Flow}}
& \multicolumn{2}{c}{\textbf{Dynamic Points}}
& \multicolumn{2}{c}{\textbf{Scene Flow}} \\
\textbf{Method}
& \textbf{Runtime (s)}
& \multicolumn{1}{c}{EPE $\downarrow$} & \multicolumn{1}{c}{APD $\uparrow$}
& \multicolumn{1}{c}{EPE $\downarrow$} & \multicolumn{1}{c}{$\threshI\uparrow$}
& \multicolumn{1}{c}{EPE $\downarrow$} & \multicolumn{1}{c}{APD $\uparrow$}
& \multicolumn{1}{c}{EPE $\downarrow$} & \multicolumn{1}{c}{$\threshI\uparrow$}
& \multicolumn{1}{c}{EPE $\downarrow$} & \multicolumn{1}{c}{APD $\uparrow$}
& \multicolumn{1}{c}{EPE $\downarrow$} & \multicolumn{1}{c}{$\threshI\uparrow$}
& \multicolumn{1}{c}{EPE $\downarrow$} & \multicolumn{1}{c}{APD $\uparrow$}
& \multicolumn{1}{c}{EPE $\downarrow$} & \multicolumn{1}{c}{$\threshI\uparrow$} \\
\midrule
MonST3R + CoTracker3     & 146.40   & 16.81   & 0.44    & 21.87    & 0.06    & 0.81   & 43.34    & 0.18    & 25.99    & 0.61   & 50.96    & 0.41    & 43.64    & 0.51   & 51.87    & 0.52    & 21.06    \\
MASt3R + CoTracker3     & 13.82   & 17.16   & 1.22    & 20.01    & 0.20    & 0.40   & 57.72    & 0.23    & 53.98    & 0.83   & 45.95    & 0.62    & 41.10    & 0.43   & 54.11    & 0.43    & 14.69    \\
VGGT + CoTracker3     & 2.31   & 8.30   & 4.80    & 11.69    & 0.77    & 0.26   & 69.12    & 0.06    & 89.37    & 0.47   & 59.21    & 0.22    & 74.11    & 0.26   & 69.34    & 0.17    & 45.77    \\
MapAnything + CoTracker3     & 0.73   & 9.42   & 2.45    & 12.88    & 0.43    & 0.25   & 70.51    & 0.06    & \textbf{89.59}    & 0.63   & 35.51    & 0.51    & 58.00    & 0.63   & 50.85    & 0.35    & \textbf{58.01}    \\
St4RTrack     & 1.12   & 11.82   & 1.03    & 14.63    & 0.10    & 0.17   & 80.87    & 0.07    & 77.90    & 0.56   & 48.11    & 0.25    & 38.31    & 0.41   & 53.12    & 0.21    & 28.46    \\
SpatialTrackerV2     & 11.56   & 5.45   & 4.48    & 10.63    & 0.10    & 0.69   & 62.34    & 0.06    & 83.66    & 0.34   & 68.37    & \textbf{0.09}    & \textbf{78.75}    & \textbf{0.21}   & \textbf{74.46}    & \textbf{0.14}    & 50.70    \\
\textbf{\coolname{}}     & \textbf{0.50}   & \textbf{3.89}   & \textbf{7.81}    & \textbf{3.14}    & \textbf{1.83}    & \textbf{0.07}   & \textbf{93.44}    & \textbf{0.05}    & 86.99    & \textbf{0.27}   & \textbf{71.70}    & 0.10    & 71.41    & 0.27   & 67.43    & 0.19    & 33.57    \\

\bottomrule
\end{tabular}%
}
\end{table*}

%% file: tables/dense_sf.tex
\begin{table*}[t]
\caption{\label{tab:dense_sf}%
  \textbf{\coolname{} achieves state-of-the-art dense scene flow estimation performance.} We report end-point error (EPE), average points within delta (APD) and inlier ratio at 0.1m ($\threshI$) for dynamic points and scene flow across three datasets, where best results are \textbf{bold}.
}
\centering
\scriptsize
\setlength\tabcolsep{4pt}
\resizebox{\linewidth}{!}{%
\begin{tabular}{lrrrrrrrrrrrr}
\toprule
& \multicolumn{4}{c}{\textbf{Kubric-4D Dynamic Camera}} & \multicolumn{4}{c}{\textbf{Kubric-4D Static Camera}} & \multicolumn{4}{c}{\textbf{VKITTI-2}} \\
\cmidrule(lr){2-5} \cmidrule(lr){6-9} \cmidrule(lr){10-13}
& \multicolumn{2}{c}{\textbf{Dynamic Points}}
& \multicolumn{2}{c}{\textbf{Scene Flow}}
& \multicolumn{2}{c}{\textbf{Dynamic Points}}
& \multicolumn{2}{c}{\textbf{Scene Flow}}
& \multicolumn{2}{c}{\textbf{Dynamic Points}}
& \multicolumn{2}{c}{\textbf{Scene Flow}} \\
\textbf{Method}
& \multicolumn{1}{c}{EPE $\downarrow$} & \multicolumn{1}{c}{APD $\uparrow$}
& \multicolumn{1}{c}{EPE $\downarrow$} & \multicolumn{1}{c}{$\threshI\uparrow$}
& \multicolumn{1}{c}{EPE $\downarrow$} & \multicolumn{1}{c}{APD $\uparrow$}
& \multicolumn{1}{c}{EPE $\downarrow$} & \multicolumn{1}{c}{$\threshI\uparrow$}
& \multicolumn{1}{c}{EPE $\downarrow$} & \multicolumn{1}{c}{APD $\uparrow$}
& \multicolumn{1}{c}{EPE $\downarrow$} & \multicolumn{1}{c}{$\threshI\uparrow$} \\
\midrule
MonST3R + SEA-RAFT     & 5.23   & 2.20    & 3.73    & 14.69    & 2.26   & 6.80    & 1.16    & 61.79    & 12.31   & 0.44    & 1.21    & 12.93    \\
MASt3R + SEA-RAFT     & 6.35   & 1.92    & 1.45    & 13.95    & 2.85   & 7.58    & 1.26    & 53.62    & 12.25   & 2.50    & 13.05    & 10.20    \\
VGGT + SEA-RAFT     & 11.80   & 3.60    & 11.76    & 14.53    & 1.92   & 15.01    & 0.78    & 86.54    & 6.57   & 2.61    & 0.70    & 37.63    \\
MapAnything + SEA-RAFT     & 17.65   & 2.67    & 17.70    & 9.16    & 2.82   & \textbf{19.99}    & 1.75    & 73.33    & 8.46   & 2.42    & 1.32    & 13.78    \\
St4RTrack     & 2.44   & 5.79    & 1.70    & 11.83    & 2.61   & 6.53    & 0.72    & 20.51    & 14.71   & 0.00    & 0.97    & 3.37    \\
\textbf{\coolname{}}     & \textbf{1.13}   & \textbf{18.14}    & \textbf{0.17}    & \textbf{83.38}    & \textbf{1.23}   & 19.53    & \textbf{0.10}    & \textbf{87.51}    & \textbf{4.97}   & \textbf{11.70}    & \textbf{0.04}    & \textbf{93.08}    \\

\bottomrule
\end{tabular}%
}
\end{table*}

%% file: tables/video_depth.tex
\begin{table}
\caption{\label{tab:video_depth}%
  \textbf{\coolname{} shows state-of-the-art video depth estimation over other single-step feed-forward baselines.} It is also competitive to iterative/optimization-based methods or ones trained specifically for this task. We report the absolute relative error ($\absrel$) and the inlier ratio at $1.25\%$ ($\delinliers$), where the best is \textbf{bold}.
}
\centering
\scriptsize
\setlength\tabcolsep{4pt}
\resizebox{\columnwidth}{!}{%
\begin{tabular}{lcccccccc}
\toprule
& \multicolumn{2}{c}{\textbf{Average}} & \multicolumn{2}{c}{\textbf{Bonn}} & \multicolumn{2}{c}{\textbf{KITTI}} & \multicolumn{2}{c}{\textbf{Sintel}} \\
\cmidrule(lr){2-3} \cmidrule(lr){4-5} \cmidrule(lr){6-7} \cmidrule(lr){8-9}
\textbf{Method}
& $\absrel\downarrow$ & $\delinliers\uparrow$
& $\absrel\downarrow$ & $\delinliers\uparrow$
& $\absrel\downarrow$ & $\delinliers\uparrow$
& $\absrel\downarrow$ & $\delinliers\uparrow$ \\
\midrule
\multicolumn{9}{l}{\textbf{a) Video Depth:}} \\[0.3em]
DepthCrafter     & \textbf{0.15}   & 85.23    & 0.07    & 97.90    & 0.11   & 88.50    & \textbf{0.27}    & \textbf{69.30}    \\
VDA     & 0.17   & \textbf{86.90}    & \textbf{0.05}    & \textbf{98.20}    & \textbf{0.08}   & \textbf{95.10}    & 0.37    & 67.40    \\

\arrayrulecolor{gray}\midrule
\multicolumn{9}{l}{\textbf{b) Feed-Forward + Iterative Optimization:}} \\[0.3em]
DUSt3R     & 0.26   & 75.83    & 0.17    & 83.50    & 0.12   & 84.90    & 0.48    & 59.10    \\
MonST3R     & 0.16   & 82.73    & 0.06    & 95.40    & 0.08   & 93.40    & 0.34    & 59.40    \\
MegaSAM     & 0.10   & 87.97    & 0.04    & 97.70    & 0.07   & 91.60    & \textbf{0.18}    & \textbf{74.60}    \\
SpatialTrackerV2     & \textbf{0.09}   & \textbf{88.80}    & \textbf{0.03}    & \textbf{98.80}    & \textbf{0.05}   & \textbf{97.30}    & 0.20    & 70.30    \\

\midrule
\multicolumn{9}{l}{\textbf{c) Single-Step Feed-Forward:}} \\[0.3em]
CUT3R     & 0.21   & 80.30    & \textbf{0.07}    & 95.00    & 0.10   & 89.90    & 0.47    & 56.00    \\
VGGT     & \textbf{0.13}   & 85.85    & \textbf{0.07}    & \textbf{97.27}    & \textbf{0.09}   & \textbf{94.37}    & \textbf{0.24}    & 65.90    \\
MapAnything     & 0.14   & 84.97    & 0.09    & 94.77    & \textbf{0.09}   & 94.26    & 0.25    & 65.87    \\
\textbf{\coolname{}}     & \textbf{0.13}   & \textbf{86.28}    & \textbf{0.07}    & \textbf{97.27}    & \textbf{0.09}   & 93.97    & \textbf{0.24}    & \textbf{67.59}    \\

\arrayrulecolor{black}\bottomrule
\end{tabular}%
}
\end{table}

%% file: tables/multi_modal.tex
\begin{table}[t]
\caption{\label{tab:multi_modal}%
  \textbf{Auxiliary inputs improve the 4D motion estimation performance of Any4D.} We compare different inputs on both dense scene flow (Kubric) and sparse 3D point tracking (LSFOdyssey) benchmarks using end-point error (EPE), average points within delta (APD) and inlier ratio at 0.1m ($\threshI$), where best is \textbf{bold}. ``Geometry" indicates use of depth, intrinsics and poses.
}
\centering
\scriptsize
\setlength\tabcolsep{4pt}
\resizebox{\columnwidth}{!}{%
\begin{tabular}{lcccccccc}
\toprule
& \multicolumn{4}{c}{\textbf{Kubric-4D Static Camera}} & \multicolumn{4}{c}{\textbf{LSFOdyssey}} \\
\cmidrule(lr){2-5} \cmidrule(lr){6-9}
& \multicolumn{2}{c}{\textbf{Dynamic Points}}
& \multicolumn{2}{c}{\textbf{Scene Flow}}
& \multicolumn{2}{c}{\textbf{Dynamic Points}}
& \multicolumn{2}{c}{\textbf{Scene Flow}} \\
\textbf{Any4D Inputs}
& EPE $\downarrow$ & APD $\uparrow$
& EPE $\downarrow$ & $\threshI\uparrow$
& EPE $\downarrow$ & APD $\uparrow$
& EPE $\downarrow$ & $\threshI\uparrow$ \\
\midrule
Images Only     & 1.17   & 21.33    & 0.11    & 86.25    & 0.28   & 71.47    & 0.12    & 68.03    \\
Images + Geometry     & \textbf{0.23}   & 80.18    & \textbf{0.09}    & 86.26    & \textbf{0.19}   & 80.80    & 0.12    & 68.71    \\
Images + Doppler     & 1.17   & 21.70    & 0.12    & 86.90    & 0.29   & 71.26    & \textbf{0.11}    & 70.32    \\
Images + Geometry + Doppler     & \textbf{0.23}   & \textbf{81.72}    & \textbf{0.09}    & \textbf{87.27}    & \textbf{0.19}   & \textbf{81.10}    & \textbf{0.11}    & \textbf{71.37}    \\

\bottomrule
\end{tabular}%
}
\end{table}

%% file: tables/representation_ablation.tex
\begin{table}[t]
\caption{\label{tab:representation_ablation}%
  \textbf{Allocentric scene flow is the optimal output representation for 4D motion.} We compare different representation types on dense scene flow (Kubric) and sparse 3D point tracking (LSFOdyssey) using end-point error (EPE), average points within delta (APD) and inlier ratio at 0.1m ($\threshI$). Best results are \textbf{bold}.
}
\centering
\scriptsize
\setlength\tabcolsep{4pt}
\resizebox{\columnwidth}{!}{%
\begin{tabular}{lcccccccc}
\toprule
& \multicolumn{4}{c}{\textbf{Kubric-4D Static Camera}} & \multicolumn{4}{c}{\textbf{LSFOdyssey}} \\
\cmidrule(lr){2-5} \cmidrule(lr){6-9}
& \multicolumn{2}{c}{\textbf{Dynamic Points}}
& \multicolumn{2}{c}{\textbf{Scene Flow}}
& \multicolumn{2}{c}{\textbf{Dynamic Points}}
& \multicolumn{2}{c}{\textbf{Scene Flow}} \\
\textbf{Representation Type}
& EPE $\downarrow$ & APD $\uparrow$
& EPE $\downarrow$ & $\threshI\uparrow$
& EPE $\downarrow$ & APD $\uparrow$
& EPE $\downarrow$ & $\threshI\uparrow$ \\
\midrule
Backprojected 2D Flow     & 2.14   & 19.44    & 1.16    & 75.69    & 0.49   & 57.21    & 0.27    & 70.11    \\
3D Points After Motion     & 1.24   & 17.33    & 0.58    & 21.84    & \textbf{0.24}   & 69.30    & 0.38    & 21.87    \\
Egocentric Scene Flow     & 1.26   & 19.43    & 0.12    & 85.37    & \textbf{0.24}   & 71.80    & 0.14    & 65.13    \\
\textbf{Allocentric Scene Flow}     & \textbf{1.23}   & \textbf{19.53}    & \textbf{0.10}    & \textbf{87.51}    & \textbf{0.24}   & \textbf{73.95}    & \textbf{0.10}    & \textbf{71.46}    \\

\bottomrule
\end{tabular}%
}
\end{table}

%% file: text/6_conclusion.tex
\section{Conclusion} 
\label{sec:conclusion}

We presented \coolname, a unified model that enables dense 4D reconstruction of dynamic scenes from both monocular and multi-modal setups. In \coolname, we chose a \emph{factorized} output representation of 4D scenes, which allowed the use of diverse data for training at scale with partial supervision for auxiliary sub-tasks, in addition to the target task of dense scene flow estimation. \coolname is flexible, and supports optional \emph{multi-modal inputs}. Importantly, we showed through our experiments that our joint training scheme produces generalizable view embeddings that improve performance whenever inputs such as depth and egocentric radial velocity (doppler) may be available to support the output prediction quantities. Finally, due to the feed-forward nature of \coolname, we saw that one can obtain dynamic scene estimates an order of magnitude faster than existing methods such as SpatialTrackerV2 \cite{xiao2024spatialtracker}, by exploiting N-view inference. We believe \coolname will ultimately enable real-time 4D scene reconstruction for applications such as Generative AI, AR/VR and Robotics, and serve as a foundational 4D reconstruction model.

%% file: supplementary.tex
\clearpage
\ifcvpr
\setcounter{page}{1}
\fi
\maketitlesupplementary

\appendix

\setcounter{section}{0}
\setcounter{equation}{0}
\setcounter{figure}{0}
\setcounter{table}{0}
\renewcommand{\thefigure}{S.\arabic{figure}}
\renewcommand{\thetable}{S.\arabic{table}}
\renewcommand{\theequation}{S.\arabic{equation}}

\ifcvpr
\begin{figure*}[!t]
\centering
\textbf{We encourage reviewers to check the supplementary website for video visualizations.}
\end{figure*}
\fi

\input{tables/datasets}

\section{Training} \label{sec:appendix_training}

\paragraph{Datasets:}
We train on a combination of static and dynamic datasets with varying levels of supervision. For supervision geometric quantities - depth, intrinsics, and camra poses, all the datasets in \ref{tab:dataset_comparison} are used. 
For scene flow supervision, we only rely on Kubric (from CoTracker3), PointOdyssey and Dynamic Replica, as they contain both diverse camera and scene motion crucial for learning good scene flow. We find that VKITTI-2 sequences span minimal scene motion while data from GCD lacks good camera diversity, and thus, only use them for geometry supervision. 

\begin{figure}[!h]
    \centering
    \includegraphics[width=\linewidth]{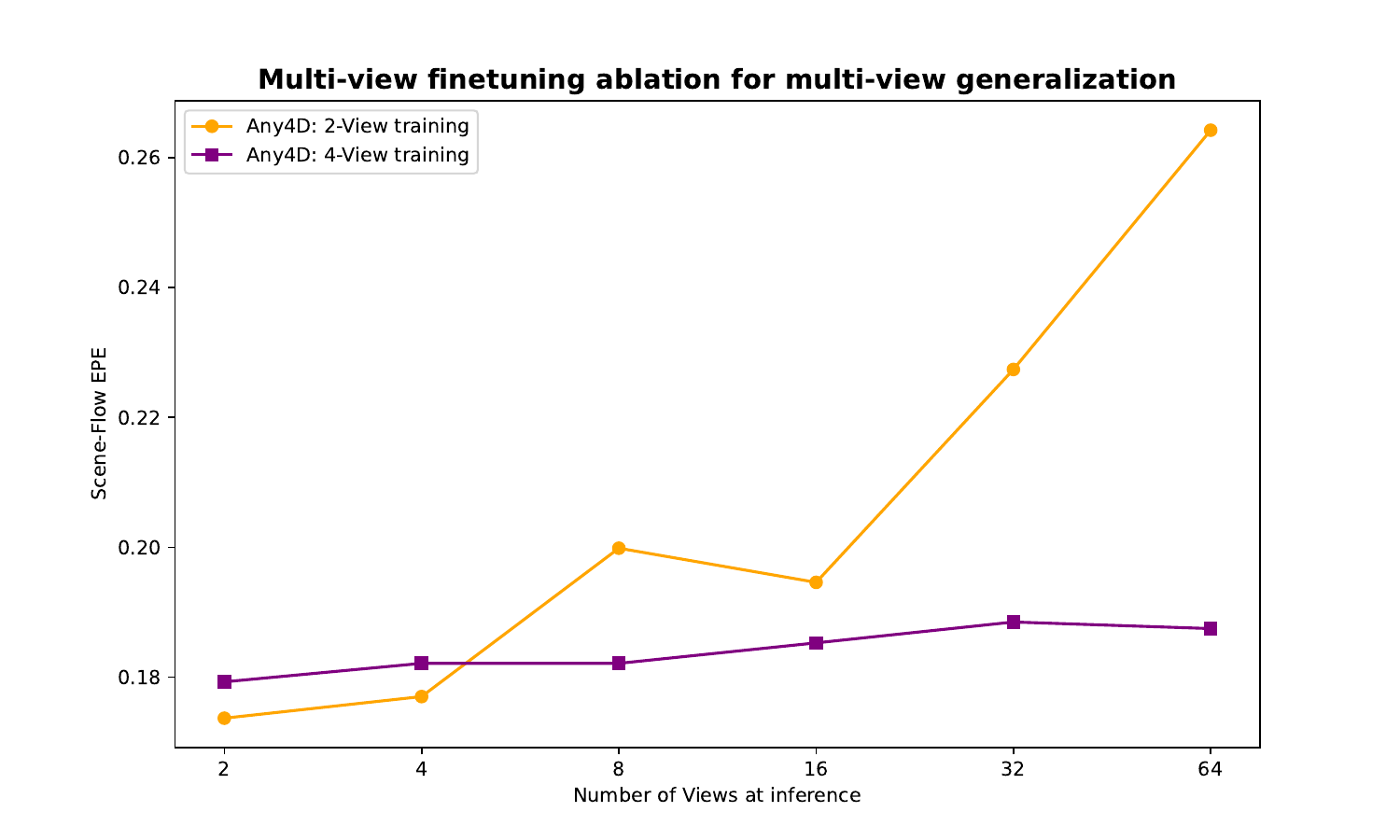}
    \caption{\textbf{4-View training is key to enabling multi-frame generalization during inference}. Any4D trained with 2 views results in higher EPE at higher number of input views. In contrast, the 4-view model exhibits stable behaviour even at 64 views. }
    \label{fig:multi_view_ablation}
\end{figure}

\paragraph{Implementation Details:}
We initialize \coolname's weights with the public MapAnything checkpoint. The doppler scene-flow encoder, and the scene-flow DPT decoder are initialized and learnt from scratch. We train the entire network with a learning rate of 1e-5, 5e-7 and 1e-4 for the entire network, the DINOv2 Image encoder and the Scene-flow DPT decoder respectively. We use a warmup of 10 epochs, and finetune the network for a total of 100 epochs, covering approximately 120k gradient steps in total on 8 H100 GPUs. The images and respective quantities in each batch cropped and resized to 518 image width, with a randomized height-width aspect ratio between 0.5 and 3. During each gradient step, we sample upto 4 views from each dataset, with a variable batch size of upto 24 views per GPU. As illustrated in \cref{fig:multi_view_ablation}, we find that 4-view training is critical for generalizing with multi-view inference.

\section{Benchmarking Setup Details}
For the TAPVID-3D PStudio dataset and DriveTrack datasets, we evaluated on a uniform subset of 50 sequences from all available datasets and use the first 64 frames for evaluation. Since the dataset is extremely sparse and each sequence only contains at most a few hundred point queries, we use all points for benchmarking. For Drive-Track, we filter 50 sequences that contain non-zero allocentric motion. For the Dynamic-Replica and LSF-Odyssey datasets, we filter out static points (i.e., points with zero allocentric motion) and use dynamic points as queries for our benchmarking, to maintain homogeneity with the 2 other datasets and emphasize benchmarking of dynamic elements of a scene. We acknowledge that our evaluation is similar to \cite{feng2025st4rtrack}.

\begin{figure}[!h]
    \centering
    \includegraphics[width=\linewidth]{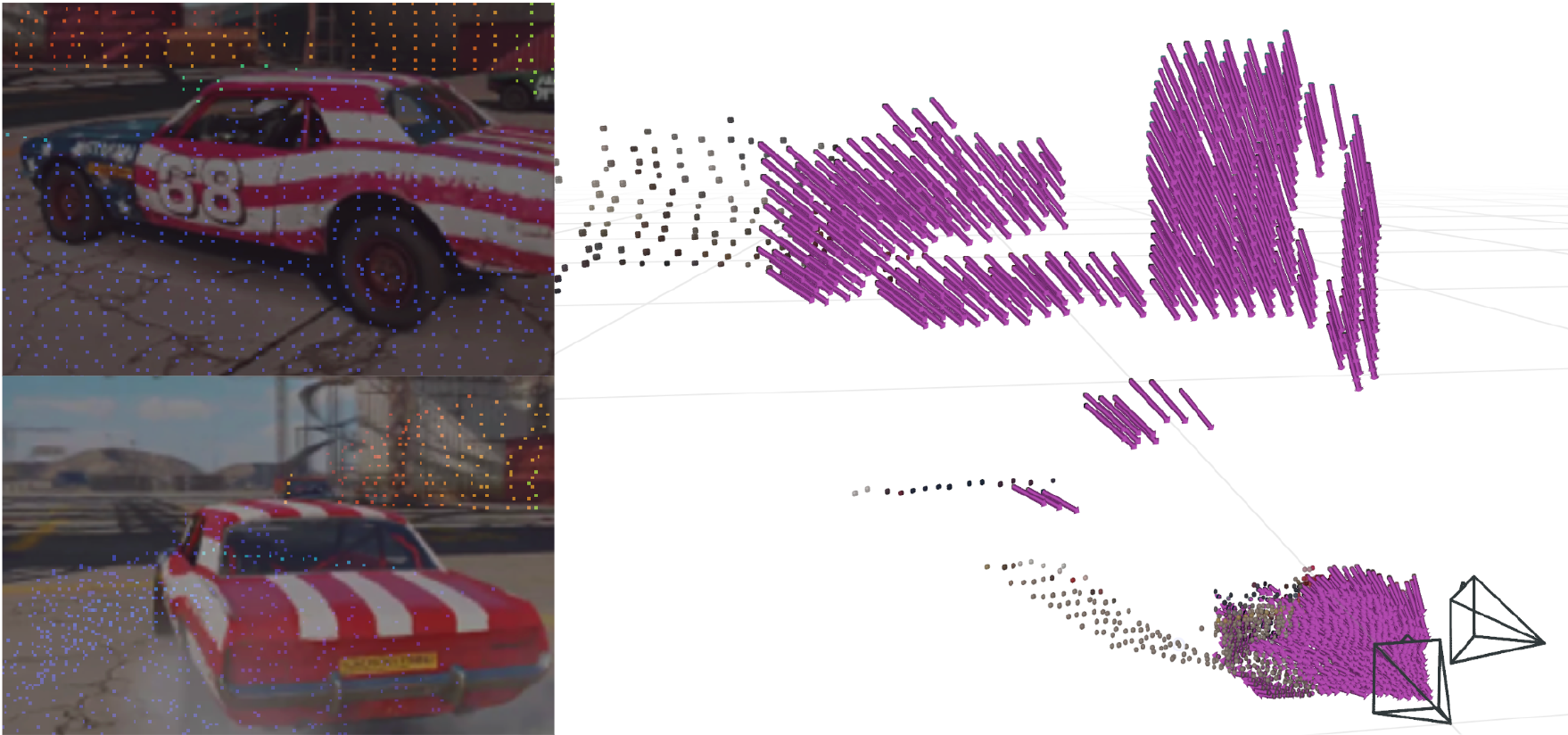}
    \caption{\textbf{Doppler Scene Flow }is simulated as radial component of ego-centric scene flow.}
    \label{fig:doppler_scene_flow}
\end{figure}

\section{Multi-Modal Conditioning}
\paragraph{Simulating Doppler Velocity:}
As shown in \cref{fig:doppler_scene_flow}, we simulate the Doppler velocity from egocentric scene flow labels. More specifically, given a 3D point $\vec{p} = [x, y, z]$ and its corresponding ego scene flow vector $\vec{v} = [\Delta x, \Delta y, \Delta z]$, the simulated Doppler velocity $v_r$ is defined as the projection of the motion vector into the radial direction of each ray. This is simply the normalized vector from the origin of the radar to the point $\vec{p}$. The Doppler (radial) velocity is computed as:

\[
v_r = \frac{\vec{p} \cdot \vec{v}}{\|\vec{p}\|} = \frac{x \cdot \Delta x + y \cdot \Delta y + z \cdot \Delta z}{\sqrt{x^2 + y^2 + z^2}}
\]

\begin{figure*}[!ht]
    \centering
    \includegraphics[width=\linewidth]{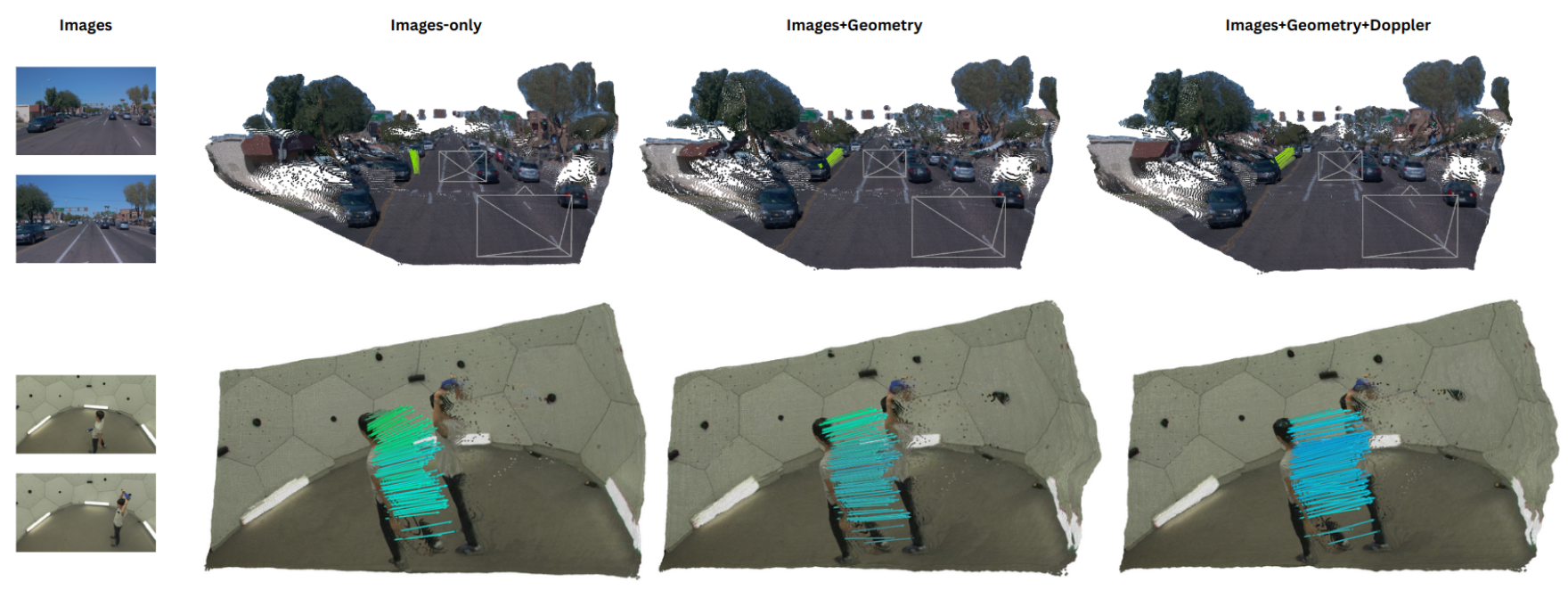}
    \caption{\label{fig:suppl_multi_modal}%
    \textbf{Qualitative visualizations of \coolname estimating 3D geometry and point tracking on TAPVID-3D Waymo Drive-Track sequences.} As visible, the image-only variant (\texttt{column 1}) sometimes produces an offset to the scene flow at the edges. However, the predictions improve whenever sparse geometry (\texttt{column 2}) and doppler annotations are available (\texttt{column 3}).
    }
\end{figure*}

\begin{figure*}[!ht]
    \centering
    \includegraphics[width=\linewidth]{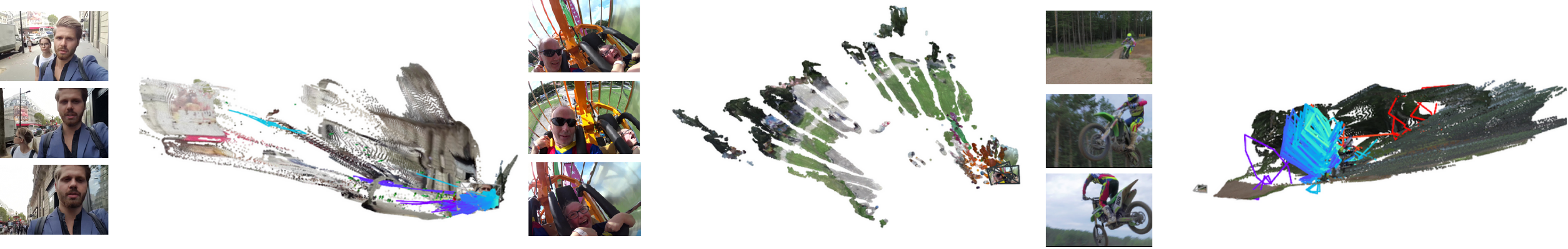}
    \caption{\label{fig:failure_cases}%
    \textbf{Qualitative visualizations of \coolname limitations.} Videos with large camera motion inducing no visual overlap of background or scene motion dominating the image space are common failure modes for Any4D. We believe that the availability of large-scale dense scene flow and 3D tracking datasets and integrating real-time optimization is key to overcoming these limitations.
    }
\end{figure*}

\ifcvprfinal
\input{text/7_acknowledgment}
\fi

\ifarxiv
\input{text/7_acknowledgment}
\fi

%% file: tables/datasets.tex
\begin{table}[!ht]
    \caption{\textbf{List of Datasets used to train \coolname}}
    \label{tab:dataset_comparison}
    \centering
    \scriptsize
    \setlength\tabcolsep{2pt}
    \begin{tabular}{lcclr}
    \toprule
    \textbf{Dataset} & \textbf{Dynamic} & \textbf{Scene Flow} & \textbf{Domain} & \textbf{\# Scenes} \\ 
    \midrule
    BlendedMVS          & \redx  & \redx & Outdoor \& object centric                    & 500 \\
    MegaDepth          & \redx  & \redx & Outdoor                    & 275 \\
    ScanNet++          & \redx  & \redx & Indoor                     & 295 \\
    VKITTI2            & \greencheck & \redx & AV                         & 40   \\
    Waymo-DriveTrack         & \greencheck & \greencheck & AV                         & 1500 \\
    GCD-Kubric             & \greencheck & \redx & Synthetic random objects    & 5000 \\
    CoTracker3-Kubric             & \greencheck & \greencheck & Synthetic random objects    & 5000 \\
   Dynamic Replica    & \greencheck & \greencheck & Synthetic humans \& animals & 500 \\
   Point Odyssey    & \greencheck & \greencheck & Diverse Synthetic assets & 159 \\

    \bottomrule
    \end{tabular}
\end{table}

%% file: text/7_acknowledgment.tex
\section*{Acknowledgments}

We thank Tarasha Khurana and Neehar Peri for their initial discussions in the project. 
We appreciate the help from Jeff Tan with setting up Stereo4D (which we ended up not using due to poor dataset quality).
Lastly, we thank Bardienus Duisterhof and members of the AirLab \& Deva's Lab at CMU for insightful discussions and feedback on the paper.

\vspace{0.4em}

\noindent This work was supported by Defense Science and Technology Agency contract {\scriptsize\texttt{\#DST000EC124000205}}, Bosch Research, and the IARPA via Department of Interior/Interior Business Center (DOI/IBC) contract {\scriptsize\texttt{140D0423C0074}}. The U.S. Government is authorized to reproduce and distribute reprints for Governmental purposes notwithstanding any copyright annotation thereon. Disclaimer: The views and conclusions contained herein are those of the authors and should not be interpreted as necessarily representing the official policies or endorsements, either expressed or implied, of IARPA, DOI/IBC, or the U.S. Government. 
Lastly, this work was supported by a hardware grant from Nvidia and used PSC Bridges-2 through allocation {\scriptsize\texttt{cis220039p}} from the Advanced Cyberinfrastructure Coordination Ecosystem: Services \& Support (ACCESS) program.